\documentclass[letterpaper,10pt,conference]{ieeeconf}
\IEEEoverridecommandlockouts
\usepackage{graphics}
\usepackage{graphicx}
\usepackage{epsfig}
\usepackage{multirow}
\usepackage{amssymb}
\usepackage{amsmath}
\usepackage{textcomp}
\usepackage{xspace}
\usepackage{makecell}
\usepackage[caption=false]{subfig}

\usepackage{amsthm}
\usepackage{mathrsfs}

\usepackage{subcaption}

\usepackage{booktabs}
\usepackage{stackengine}
\usepackage{xfrac}
\usepackage{tabularx}
\usepackage{multirow}
\usepackage{cite}
\usepackage{lipsum}
\usepackage{cancel}
\usepackage[normalem]{ulem} 
\usepackage{bm}
\usepackage{soul} 
\usepackage{stmaryrd} 

\usepackage[linesnumbered,ruled,vlined]{algorithm2e}
\let\labelindent\relax
\usepackage{enumitem}

\usepackage[integrals]{wasysym}

\usepackage{threeparttable}
\usepackage{optidef}
\def\matt#1{\begin{bmatrix}#1\end{bmatrix}}

\makeatletter
\let\NAT@parse\undefined

\makeatother
\usepackage[hidelinks]{hyperref}
\hypersetup{
    colorlinks=true,
    linkcolor=blue,
    filecolor=magenta,
    urlcolor=cyan,
    citecolor=green
}

\usepackage[dvipsnames]{xcolor,colortbl}

\newtheorem{lemma}{Lemma}

\newtheorem{remark}{Remark}

\begin{document}

\title{
Online Intention Prediction via Control-Informed Learning
}

\author{
Tianyu Zhou$^{1}$, Zihao Liang$^{2}$, Zehui Lu$^{1}$ and Shaoshuai Mou$^{1}$
\thanks{This material is based upon work supported by the Office of Naval Research (ONR) and Saab, Inc. under the Threat and Situational Understanding of Networked Online Machine Intelligence (TSUNOMI) program (grant no. N00014-23-C-1016). Any opinions, ﬁndings and conclusions or recommendations expressed in this material are those of the author(s) and do not necessarily reﬂect the views of the ONR, the U.S. Government, or Saab, Inc.”}
\thanks{$^{1}$T. Zhou, Z. Lu and S. Mou are with the School of Aeronautics and Astronautics, Purdue University, IN 47907, USA {\tt\small \{zhou1043, lu846, mous\}@purdue.edu}}
\thanks{$^{2}$Z. Liang is an independent researcher {\tt\small danlzh@outlook.com}}
}

\maketitle

\begin{abstract}
This paper presents an online intention prediction framework for estimating the goal state of autonomous systems in real time, even when intention is time-varying, and system dynamics or objectives include unknown parameters. The problem is formulated as an inverse optimal control / inverse reinforcement learning task, with the intention treated as a parameter in the objective. A shifting horizon strategy discounts outdated information, while online control-informed learning enables efficient gradient computation and online parameter updates. Simulations under varying noise levels and hardware experiments on a quadrotor drone demonstrate that the proposed approach achieves accurate, adaptive intention prediction in complex environments.
\end{abstract}

\section{Introduction}

Intention prediction refers to the process of inferring the future goals or desired outcomes of an autonomous agent or human based on its observed behavior or state trajectory \cite{mcghan2015human,liu2020spatiotemporal}.
It is crucial in numerous applications, such as human-robot interaction and collaborative systems, where anticipating future actions allows systems to react more effectively and safely, potentially without communication. In this work, we focus on cases where an agent’s intention is represented as its goal state, so intention prediction is formulated as estimating the underlying goal state from observed trajectories.

Approaches to intention prediction generally fall into two categories: physics-based and learning-based.
Physics-based methods exploit dynamics or kinematics models to forecast trajectories with low computational cost but depend on accurate dynamics, which are often hard to obtain in real-world settings with disturbances or nonlinearities \cite{rudenko2020human,huang2022survey}. Examples include single-trajectory models \cite{brannstrom2010model}, Kalman filters \cite{lefkopoulos2020interaction}, and Monte Carlo sampling \cite{wang2019trajectory}. Learning-based methods use data-driven models to capture nonlinear or unmodeled effects. Classic approaches include Gaussian processes, support vector machines, hidden Markov models, and dynamic Bayesian networks \cite{guo2019modeling,kumar2013learning,li2020pedestrian}, while recent work applies deep and generative models \cite{chandra2020forecasting,gu2021densetnt}. These methods can generalize poorly outside the training distribution and often incur high computational cost, limiting real-time use \cite{rudenko2020human,huang2022survey}.

\begin{figure}[t!]
\centering\includegraphics[width=0.8\linewidth]{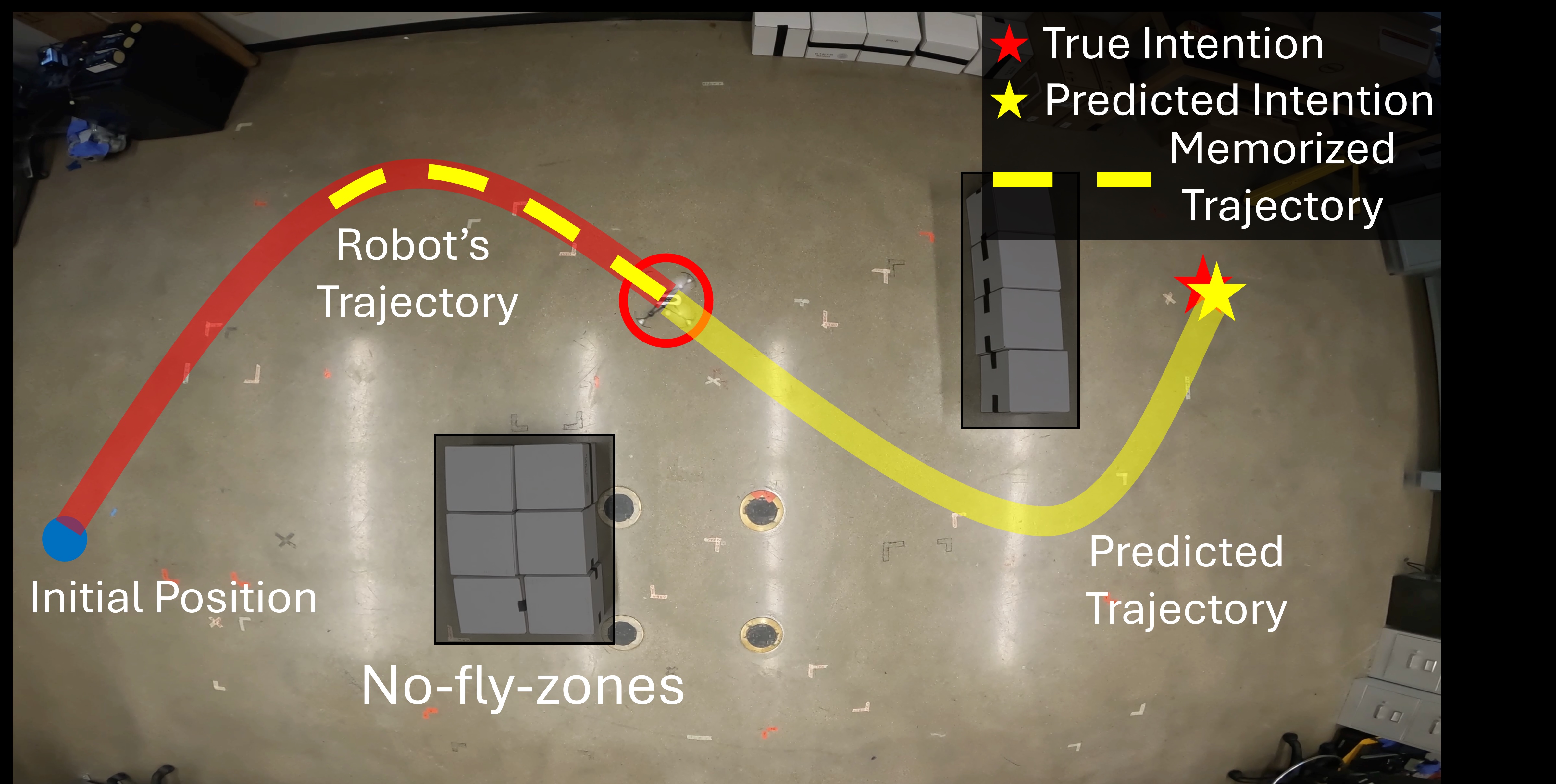}
\caption{A robot with an unknown intention was observed while executing its trajectory autonomously, during which the proposed algorithm predicted its intention in real time.} \label{fig:snapshot}
\end{figure}

Among learning-based approaches, imitation learning is a widely used framework for intention prediction. It includes both inverse reinforcement learning (IRL) and inverse optimal control (IOC) \cite{wang2019imitation,macglashan2015between}. The goal is to recover an unknown reward or objective from demonstrations, often written as a weighted sum of features \cite{abbeel2004apprenticeship,ziebart2008maximum,liang2022iterative}. With system dynamics, the estimated objective is then minimized to predict future trajectories. Discrepancies between these predictions and observed behavior are then used to update the estimate, forming a loop that improves intention prediction over time \cite{wu2020efficient,huang2021driving}.

Despite notable advances in intention prediction, current methods do not adequately address scenarios where an autonomous system’s goal may change during execution. Most studies focus on predicting a single goal or selecting from a predefined and fixed set of candidate goals, which limits their applicability in adversarial or highly dynamic environments \cite{gu2021densetnt,zhao2021tnt}. Physics-based approaches are computationally efficient but depend on accurate models of dynamics and objectives, so they cannot cope well with unknown parameters. Imitation learning-based methods can infer objectives from demonstrations but are typically designed for offline settings and cannot update estimates as new data arrive \cite{hu2022model,duan2024structured}. Online prediction is essential, as a system’s goals and environment can change rapidly, requiring continuous updates for responsive prediction. Few existing approaches jointly handle changing goals and unknown parameters in both the objective and the dynamics while also performing online estimation, which are critical capabilities for accurate intention prediction in real-world applications. Methods such as Pontryagin Differentiable Programming (PDP) \cite{jin2020pontryagin} and data-driven IOC \cite{liang2023data} are formulated for offline learning and therefore do not support online estimation of objective or dynamics parameters. To address this limitation, Online Control-Informed Learning (OCIL) combines physics-based models with learning-based components by leveraging optimal control principles and online state estimation to enable real-time goal inference from streaming observations \cite{liang2024online,zhou2025safe}. However, OCIL is primarily designed for single-goal settings and does not generalize well to scenarios involving changing or multiple goals.

Our contributions are as follows. We formulate a new intention prediction problem that permits goals to change during execution (time-varying goals), motivated by the behavior of adversarial or highly dynamic autonomous systems. Unlike prior work that presumes a fixed or discrete set of goals, our formulation estimates the goal in a continuous space and accommodates unknowns in both the objective function and the system dynamics, overcoming limitations of physics-based approaches that rely on precise models. To solve this problem, we: (i) cast intention prediction as an IOC/IRL problem, treating the goal state as a parameter in the objective rather than relying on a predefined goal set; (ii) propose a shifting horizon intention prediction strategy that discounts outdated information to better manage time-varying intention; and (iii) identify the goal online at each time step as new noisy measurements become available. 

\textbf{Notations.} 
$\lVert \cdot \lVert$ denotes the Euclidean norm.
For positive integers $n$ and $m$, let $\boldsymbol{I}_{n}$ be the ${n \times n}$ identity matrix; $\mathbf{0}_n \in\mathbb{R}^n$ denotes a vector with all zeros; $\mathbf{0}_{n\times m}$ denotes a $n\times m$ matrix with all zeros. Let $\text{col}\{\boldsymbol{v}_1,\hdots,\boldsymbol{v}_a\}\triangleq[\boldsymbol{v}_1^\prime \hdots \boldsymbol{v}_a^\prime]^\prime$ denote a column stack of elements $\boldsymbol{v}_1,\hdots,\boldsymbol{v}_a$.
Let $\mathcal{N}(\boldsymbol{a},\boldsymbol{B})$ denote a multivariate Gaussian distribution, where $\boldsymbol{a}\in \mathbb{R}^m$ is the mean and $\boldsymbol{B}\in \mathbb{R}^{m\times m}$ is the covariance matrix.
For a vector $\boldsymbol{v}$, $\boldsymbol{v}[a:b]$ slices a subvector from the $a$-th until $b-$th element, with both ends included; $\boldsymbol{v}[1]$ is the first element.

\section{Problem Formulation} \label{section:PF}

This paper addresses the intention prediction problem by predicting the goal state of a class of autonomous systems whose behavior is governed by the minimization of a control objective function. 
The system is assumed to have \emph{unknown} time-varying goal states $\boldsymbol{x}_{\mathrm{g},t}^*\in\mathbb{R}^n$ for time index $t=0,1,\ldots,T$ with $T$ the final time. In other words, during trajectory execution, the system may switch from one goal to another goal state at unknown times. 
In addition, the system contains a set of \emph{unknown} parameters that appear both in the control objective and in the system dynamics. 
Prior to any goal switching, the trajectory of the autonomous system is obtained by solving the following optimal control problem:
\begin{subequations} \label{eq:auto_sys}
\begin{align}
&\{\boldsymbol{x}_{1:T}^*,\boldsymbol{u}_{0:T-1}^*\} 
= \arg \min_{\boldsymbol{x}_{1:T},\boldsymbol{u}_{0:T-1}} 
J, 
\label{eq:obj}\\
&\text{s.t. } 
\boldsymbol{x}_{t+1} = \boldsymbol{f}(\boldsymbol{x}_t, \boldsymbol{u}_t, \boldsymbol{p}^*),
\quad \text{with}\quad \boldsymbol{x}_0 \text{ given}, 
\label{eq:dyn}
\end{align} \label{eq:opt_sys} 
\end{subequations}
where $J = \sum_{t=0}^{T-1}{\boldsymbol{\omega}_{\mathrm{r}}^*}^{\prime}
c_{\mathrm{r}}(\boldsymbol{x}_t,\boldsymbol{u}_t,\boldsymbol{x}_{\mathrm{g},0}^*)
+{\boldsymbol{\omega}_{\mathrm{f}}^*}^{\prime} 
c_{\mathrm{f}}(\boldsymbol{x}_T,\boldsymbol{x}_{\mathrm{g},0}^*)$.
The state and control input are 
\(\boldsymbol{x}_t\in\mathbb{R}^n\) and 
\(\boldsymbol{u}_t\in\mathbb{R}^m\); 
\(\boldsymbol{x}_{1:T}^*\) and \(\boldsymbol{u}_{0:T-1}^*\) denote the column stacks of optimal states and control inputs. 
The unknown weight vectors 
\(\boldsymbol{\omega}_{\mathrm{r}}^*\in\mathbb{R}^r\) and 
\(\boldsymbol{\omega}_{\mathrm{f}}^*\in\mathbb{R}^q\) appear in the running and final cost feature functions 
\(c_{\mathrm{r}}:\mathbb{R}^n\times\mathbb{R}^m\times\mathbb{R}^n\mapsto\mathbb{R}^r\) 
and 
\(c_{\mathrm{f}}:\mathbb{R}^n\times\mathbb{R}^n\mapsto\mathbb{R}^q\), both assumed differentiable. 
The system dynamics \eqref{eq:dyn} are parameterized by an unknown constant vector 
\(\boldsymbol{p}^*\in\mathbb{R}^p\), with 
\(\boldsymbol{f}:\mathbb{R}^n\times\mathbb{R}^m\times\mathbb{R}^p\mapsto\mathbb{R}^n\) 
twice differentiable.

Each time the autonomous system switches its goal state, i.e., at time \(\bar{t}\), the remaining trajectory from 
\(\bar{t}\) to the final time \(T\) is recomputed as
\begin{subequations}
\begin{align}
&\{\boldsymbol{x}_{\bar{t}+1:T}^*,\boldsymbol{u}_{\bar{t}:T-1}^*\} = \arg \min_{\boldsymbol{x}_{\bar{t}+1:T}^*,\boldsymbol{u}_{\bar{t}:T-1}^*} \bar{J},\\
&\text{s.t. } \boldsymbol{x}_{t+1} = \boldsymbol{f}(\boldsymbol{x}_t, \boldsymbol{u}_t, \boldsymbol{p}^*), \  \text{with}\ \boldsymbol{x}_{\bar{t}}^*\ \text{as initial state},
\end{align}  \label{eq:new_traj}
\end{subequations}
where $\bar{J} = \sum_{t=\bar{t}}^{T-1}{\boldsymbol{\omega}_{\mathrm{r}}^*}^{\prime}
c_{\mathrm{r}}(\boldsymbol{x}_t,\boldsymbol{u}_t,\boldsymbol{x}_{\mathrm{g},\bar{t}}^*)
+{\boldsymbol{\omega}_{\mathrm{f}}^*}^{\prime} 
c_{\mathrm{f}}(\boldsymbol{x}_T,\boldsymbol{x}_{\mathrm{g},\bar{t}}^*)$.
At each time \(t\), a noisy observation $\bar{\boldsymbol{x}}_t^*=\boldsymbol{x}_t^*+\boldsymbol{v}_t \in \mathbb{R}^n$ is obtained,
where \(\boldsymbol{v}_t \sim \mathcal{N}(\boldsymbol{0}_{n},\boldsymbol{R}_t)\) 
is zero-mean multivariate Gaussian measurement noise with covariance 
\(\boldsymbol{R}_t\in\mathbb{R}^{n\times n}\), assumed to be small.
In the present work, we assume full-state measurements with small noise, which allows us to use the measurement to propagate the prediction. A direction for future work is to extend the framework to partial or nonlinear output measurements under appropriate observability conditions.
Let \(\hat{\boldsymbol{x}}_{\mathrm{g},t}\in\mathbb{R}^n\) be the estimation of 
\(\boldsymbol{x}_{\mathrm{g},t}^*\) at time \(t\). 
Intention prediction is achieved by estimating this unknown goal state 
\(\boldsymbol{x}_{\mathrm{g},t}^*\) from observations.

With system \eqref{eq:opt_sys}, given its noisy observation $\bar{\boldsymbol{x}}_t^*$ at every time $t$, \textbf{the problem of interest} is to develop an online method that updates the estimation $\hat{\boldsymbol{x}}_{\mathrm{g},t}$, such that $||\hat{\boldsymbol{x}}_{\mathrm{g},t} - \boldsymbol{x}_{\mathrm{g},t}^*||^2\rightarrow 0 $ as $t\rightarrow T$.

\begin{remark}
This problem differs from existing work in the literature in three key aspects. 
First, it explicitly considers an autonomous system with unknown parameters in both system dynamics and control objectives, which will be estimated with intention in real-time. 
Second, it accounts for time-varying goal states, a factor that poses a significant challenge for existing intention prediction methods. Third, it estimates the goal directly in continuous space, while also addressing the more challenging case where no predefined, finite set of possible goals is available \cite{elnaggar2018irl,best2015bayesian,zhao2021tnt,gu2021densetnt}.
\end{remark}

\section{Proposed Algorithm} \label{section:method}
This section presents the proposed Online Intention Prediction algorithm. To accommodate switching intentions, we first introduce a shifting horizon strategy that retains only a limited window of past observations, allowing the predictor to discount outdated information and adapt to goal changes. Within this horizon, the algorithm performs an online parameter update, simultaneously estimating the unknown goal state and other system parameters. The required gradients for this update are computed using PDP \cite{jin2020pontryagin}, enabling efficient and accurate real-time estimation.

\subsection{Shifting Horizon Intention Prediction}

Online intention prediction is performed by forecasting the system’s future trajectory using estimated parameters online. 
Rather than optimizing over the entire history, i.e., start from $\boldsymbol{x}_0^*$, we restrict the computation to a recent horizon of the trajectory. 
This shifting horizon has two main advantages. 
First, for time-varying goals, older data can correspond to a different objective; limiting the window keeps the estimate aligned with the system’s current intention. 
Second, it improves robustness to noise and nonstationarity, as measurement noise and parameter drift accumulate over time; a sliding window mitigates these effects.

\begin{figure}
\centering\includegraphics[width=\linewidth]{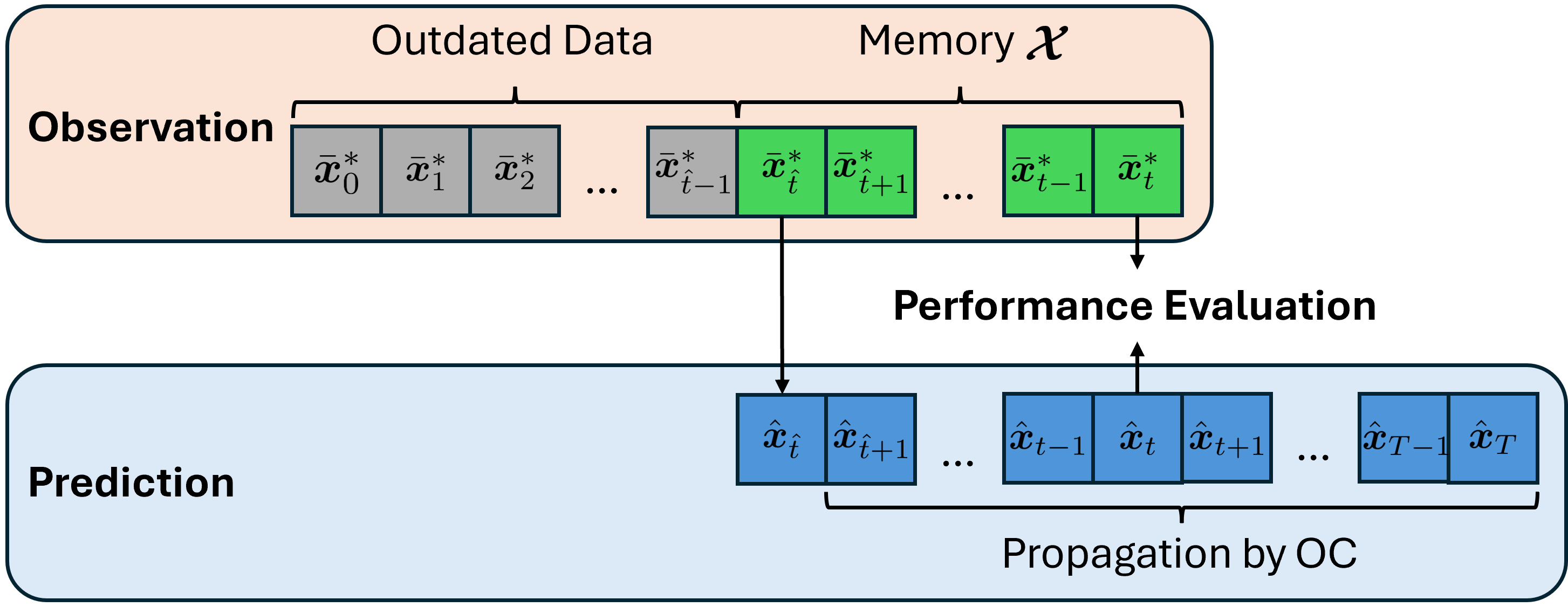}
\caption{Strategy of shifting horizon in prediction. 
} \label{fig:explain}
\end{figure}

To formalize this idea, we introduce a \emph{memory time} \(T_{\mathrm{m}}\leq T\), which defines a memory buffer 
\(\boldsymbol{\mathcal{X}}=\{\bar{\boldsymbol{x}}_{\hat{t}}^*,\ldots,\bar{\boldsymbol{x}}_{t}^*\}\) that stores the observations. 
The starting index of this buffer is 
\begin{equation}
    \hat{t}=\max(t-T_{\mathrm{m}},0),
\end{equation}
where \(t\) is the current time.
Choosing \(T_{\mathrm{m}}\) achieves a balance between \emph{stability}, which means retaining enough data to filter noise, and \emph{adaptability}, which means discarding obsolete information so the estimate tracks the current intention.

At each time step \(t\), the predicted trajectory is obtained by solving the following optimal control problem:
\begin{subequations}\label{eq:est_auto_sys}
\begin{align}
& \{\hat{\boldsymbol{x}}_{\hat{t}+1:T}, \hat{\boldsymbol{u}}_{\hat{t}:T-1}\} 
= \arg\min_{\boldsymbol{x}_{\hat{t}+1:T},\boldsymbol{u}_{\hat{t}:T-1}} 
\hat{J}, 
\label{eq:est_obj} \\
& \text{s.t. } \boldsymbol{x}_{\tau+1} 
= \boldsymbol{f}(\boldsymbol{x}_\tau, \boldsymbol{u}_\tau, \hat{\boldsymbol{p}}_t),
\quad 
\text{with}\quad \hat{\boldsymbol{x}}_{\hat{t}}=\bar{\boldsymbol{x}}_{\hat{t}}^*, 
\label{eq:est_dyn}
\end{align}
\end{subequations}
where $\hat{J}
= \sum_{\tau=\hat{t}}^{T-1} 
\hat{\boldsymbol{\omega}}_{\mathrm{r},t}^{\prime} 
c_{\mathrm{r}}(\boldsymbol{x}_\tau,\boldsymbol{u}_\tau,\hat{\boldsymbol{x}}_{\mathrm{g},t})
+\hat{\boldsymbol{\omega}}_{\mathrm{f},t}^{\prime} 
c_{\mathrm{f}}(\boldsymbol{x}_T,\hat{\boldsymbol{x}}_{\mathrm{g},t})$.
Fig.~\ref{fig:explain} illustrates the process of shifting horizon strategy. 
At current time \(t\), the predicted trajectory from \eqref{eq:est_auto_sys} is propagated from a previously observed state \(\bar{\boldsymbol{x}}_{\hat{t}}^*\), located \(T_{\mathrm{m}}\) steps earlier, using the current estimates of the goal state \(\hat{\boldsymbol{x}}_{\mathrm{g},t}\) and system parameters. 
If insufficient observations are available, i.e., when \(t<T_{\mathrm{m}}\), the trajectory is instead propagated from the initial state \(\boldsymbol{x}_0^*\). 
The discrepancy between the current observed state \(\bar{\boldsymbol{x}}_t^*\) and the predicted state at time \(t\), \(\hat{\boldsymbol{x}}_t\), will be used to evaluate prediction performance, as described later.

\begin{remark}
Accurate intention prediction requires estimating the unknown parameters 
\(\boldsymbol{\omega}_r^*,\boldsymbol{\omega}_f^*\) and \(\boldsymbol{p}^*\). 
In an optimal control system, these weights encode the agent’s objectives and trade-offs, directly shaping its trajectory. 
Recovering them is therefore essential: without identifying the parameters governing the cost and dynamics, the same trajectory could arise from different goals, making the intention ambiguous.
\end{remark}

The goal state $\boldsymbol{x}_{\mathrm{g},t}^*$ is treated as an unknown parameter in the system. For notation simplicity, define time-varying parameters $\boldsymbol{\theta}_t^*\triangleq \text{col}\{\boldsymbol{p}^*,\boldsymbol{\omega}_{\mathrm{r}}^*,\boldsymbol{\omega}_{\mathrm{f}}^*,\boldsymbol{x}_{\mathrm{g},t}^*\}\in\mathbb{R}^{s}$  where $s=p+r+q+n$, and its estimation at time $t$ to be $\hat{\boldsymbol{\theta}}_t$.

\subsection{Online Parameter Estimation}

To obtain $\boldsymbol{\theta}_t^*$, the parameter estimation problem is converted into a state estimation problem, where the state of the new system is $\boldsymbol{\theta}$. The new system dynamics and measurement are defined as follows:
\begin{subequations} \label{eq:dynTheta}
\begin{align}
\boldsymbol{\theta}_{t} &= 
\boldsymbol{\theta}_t^* \text{ (unknown)} \label{eq:dynTheta:1} 
, 
\\
\bar{\boldsymbol{x}}_t^*&=\boldsymbol{x}_t^*+\boldsymbol{v}_t, \label{eq:dynTheta:2}
\end{align}
\end{subequations}
where \eqref{eq:dynTheta:1} indicates the new dynamics and \eqref{eq:dynTheta:2} represents the new measurement. 
The timing and magnitude of parameter changes are unknown and must be inferred from observed changes in the system’s behavior. 
This inference is possible because the system’s trajectory is simultaneously reinitialized at the change point. The influence of the previous goal may persist for a short period after a change, but it gradually diminishes as the prediction horizon shifts.
The performance of the estimation at time $t>0$ is evaluated using the following residual function:
\begin{equation} \label{eq:residual}
\boldsymbol{l}(\hat{\boldsymbol{x}}_t(\hat{\boldsymbol{\theta}}_{t-1}),\bar{\boldsymbol{x}}_t^*)=\bar{\boldsymbol{x}}_t^* - \hat{\boldsymbol{x}}_t(\hat{\boldsymbol{\theta}}_{t-1})\in \mathbb{R}^n,
\end{equation}
where $\hat{\boldsymbol{x}}_t(\hat{\boldsymbol{\theta}}_{t-1})$ denotes the estimated state at time $t$ by solving \eqref{eq:est_auto_sys} with the latest estimated parameter $\hat{\boldsymbol{\theta}}_{t-1}$. 
The online parameter estimation is achieved by implementing \cite{liang2024online} with the following lemma:
\begin{lemma}\label{lemma:OCIL}
(Online Control-Informed Learning \cite{liang2024online}) Let $\boldsymbol{H}_t = \frac{d \boldsymbol{l}(\hat{\boldsymbol{x}}_t(\boldsymbol{\hat{\theta}}_{t-1}),\bar{\boldsymbol{x}}_t^*)}{d\hat{\boldsymbol{\theta}}_{t-1}} \in \mathbb{R}^{n\times s}$ be a uniformly bounded and full rank matrix for every $\boldsymbol{\theta}$, 
and introduce unknown diagonal matrices $\boldsymbol{\mathcal{F}}_{t}$ and $\boldsymbol{\mathcal{G}}_{t}$ to model the measurement and prediction error \cite{boutayeb2002convergence}.
If the following inequalities hold: $(\boldsymbol{\mathcal{F}}_t-\boldsymbol{I}_{s})^2\le\boldsymbol{R}_t(\boldsymbol{H}_t\boldsymbol{P}_{t|t-1}\boldsymbol{H}^\prime_t+\boldsymbol{R}_t)^{-1}$, $\boldsymbol{\mathcal{G}}_{t}^{\prime}\boldsymbol{P}_{t|t-1}^{-1}\boldsymbol{\mathcal{G}}_{t}-\boldsymbol{P}_{t|t-1}^{-1}\leq 0$,
then the parameter estimator:
\begin{subequations} \label{eq:EKF}
\begin{align}
\text{Predict: }& \hat{\boldsymbol{\theta}}_{t|t-1} = \hat{\boldsymbol{\theta}}_{t-1}, \ \boldsymbol{P}_{t|t-1} = \boldsymbol{P}_{t-1}, \label{eq:EKF:theta} \\
\text{Update: }& \boldsymbol{K}_t = \boldsymbol{P}_{t|t-1}\boldsymbol{H}_t'(\boldsymbol{H}_t \boldsymbol{P}_{t|t-1} \boldsymbol{H}_t' + \boldsymbol{R}_t)^{-1}, \label{eq:EKF:K} \\
& \boldsymbol{P}_{t} = (\boldsymbol{I}_s - \boldsymbol{K}_t \boldsymbol{H}_t)\boldsymbol{P}_{t|t-1}, \label{eq:EKF:P} \\
& \hat{\boldsymbol{\theta}}_t = \hat{\boldsymbol{\theta}}_{t|t-1} - \boldsymbol{K}_t (\bar{\boldsymbol{x}}_t^*-\hat{\boldsymbol{x}}_t(\hat{\boldsymbol{\theta}}_t)). \label{eq:EKF:theta_final}
\end{align}
\end{subequations}
ensures the local asymptotic convergence of $\hat{\boldsymbol{\theta}}$ to $\boldsymbol{\theta}^*$. The subscript $_{t|t-1}$ represents the quantity that has not yet been updated at time $t$; $\boldsymbol{P}_t\in\mathbb{R}^{s\times s}$ is a positive-definite matrix that denotes the covariance of the estimation; $\boldsymbol{K}_t\in\mathbb{R}^{s \times n}$ denotes the optimal prediction gain.
\end{lemma}
\begin{remark}
    Different from OCIL \cite{liang2024online}, which assumes fixed parameters, this paper considers time-varying parameters \(\boldsymbol{\theta}_t^*\). The shifting-horizon strategy improves the estimation of time-varying parameters, which OCIL cannot handle effectively, as demonstrated in the next section.
\end{remark}
Everything in \eqref{eq:EKF} is known except $\boldsymbol{H}_t$. For notation simplicity, $\frac{d \boldsymbol{l}(\hat{\boldsymbol{x}}_t(\hat{\boldsymbol{\theta}}_{t-1}))}{d \hat{\boldsymbol{\theta}}_{t-1}}$ is used to represent $\frac{d \boldsymbol{l}(\hat{\boldsymbol{x}}_t(\boldsymbol{\hat{\theta}}_{t-1}),\bar{\boldsymbol{x}}_t^*)}{d\hat{\boldsymbol{\theta}}_{t-1}}$.
Applying the chain rule, we have
\begin{equation} \label{eq:chainRule}
\boldsymbol{H}_t=
\frac{d \boldsymbol{l}(\hat{\boldsymbol{x}}_t(\hat{\boldsymbol{\theta}}_{t-1}))}{d \hat{\boldsymbol{\theta}}_{t-1}}=\frac{\partial \boldsymbol{l}(\hat{\boldsymbol{x}}_t(\hat{\boldsymbol{\theta}}_{t-1}))}{\partial \hat{\boldsymbol{x}}_t(\hat{\boldsymbol{\theta}}_{t-1})}\frac{\partial  \hat{\boldsymbol{x}}_t(\hat{\boldsymbol{\theta}}_{t-1})}{\partial \hat{\boldsymbol{\theta}}_{t-1}},
\end{equation}
where $\frac{\partial \boldsymbol{l}(\hat{\boldsymbol{x}}_t(\hat{\boldsymbol{\theta}}_{t-1}))}{\partial \hat{\boldsymbol{x}}_t(\hat{\boldsymbol{\theta}}_{t-1})}$ is a known gradient of the residual function with respect to the state at time $t$, evaluated at $\hat{\boldsymbol{\theta}}_{t-1}$; and $\frac{\partial  \hat{\boldsymbol{x}}_t(\hat{\boldsymbol{\theta}}_{t-1})}{\partial \hat{\boldsymbol{\theta}}_{t-1}}$ is an unknown gradient of the state at time $t$ with respect to $\boldsymbol{\theta}$, evaluated at $\hat{\boldsymbol{\theta}}_{t-1}$.
\cite{jin2020pontryagin} introduces a method to obtain the gradient $\frac{\partial  \hat{\boldsymbol{x}}_t(\hat{\boldsymbol{\theta}}_{t-1})}{\partial \hat{\boldsymbol{\theta}}_{t-1}}$, as provided in the following.

\begin{lemma}[PDP \cite{jin2020pontryagin}]\label{lemma:PDP}
Given an arbitrary system from \eqref{eq:est_auto_sys}, its Hamiltonian is formulated as, $\boldsymbol{\mathcal{H}}_{t} = \boldsymbol{\omega}_{\mathrm{r}}^{\prime}c_{\mathrm{r}}(\boldsymbol{x}_{t},\boldsymbol{u}_{t},\boldsymbol{x}_{\mathrm{g}})+\boldsymbol{f}(\boldsymbol{x}_{t},\boldsymbol{u}_{t},\boldsymbol{p})^\prime\boldsymbol{\lambda}_{t+1}$, $\boldsymbol{\mathcal{H}}_T=\boldsymbol{\omega}_{\mathrm{f}}'c_{\mathrm{f}}(\boldsymbol{x}_T,\boldsymbol{x}_{\mathrm{g}})$,
for all $t=\hat{t},\cdots,T-1$, where $\boldsymbol{\lambda}_{t}\in\mathbb{R}^n$ denotes the costate and $\boldsymbol{x}_{\mathrm{g}}$ denotes the goal at $t$. Let $\boldsymbol{\mathcal{H}}_{xx,t}$ denote the second-order derivative of $\boldsymbol{\mathcal{H}}_t$ with respect to $\boldsymbol{x}$, and similar notations for $\boldsymbol{\mathcal{H}}_{uu,t}$, $\boldsymbol{\mathcal{H}}_{xu,t}=\boldsymbol{\mathcal{H}}_{ux,t}'$, $\boldsymbol{\mathcal{H}}_{x\theta,t}$,  $\boldsymbol{\mathcal{H}}_{u\theta,t}$, $\boldsymbol{\mathcal{H}}_{xx,T}$, and $\boldsymbol{\mathcal{H}}_{x\theta,T}$. Let $\boldsymbol{F}_t,\boldsymbol{G}_t,\boldsymbol{E}_t$ denote the first-order derivatives of dynamics.
If $\boldsymbol{\mathcal{H}}_{uu,t}$ is invertible for all $t=T-1,\cdots,\hat{t}$, we have:
\begin{equation*} \label{eq:backwardPass}
\begin{aligned}
&\boldsymbol{V}_{t}=\boldsymbol{C}_{t}+\boldsymbol{A}_{t}^\prime(\boldsymbol{I}+\boldsymbol{V}_{t+1}\boldsymbol{B}_{t})^{-1}\boldsymbol{V}_{t+1}\boldsymbol{A}_{t}, \\&\boldsymbol{W}_{t}=\boldsymbol{A}_{t}^\prime(\boldsymbol{I}+\boldsymbol{V}_{t+1}\boldsymbol{B}_{t})^{-1}(\boldsymbol{W}_{t+1}+\boldsymbol{V}_{t+1}\boldsymbol{M}_{t})+\boldsymbol{N}_{t},
\end{aligned}
\end{equation*}
with $\boldsymbol{V}_{T}=\boldsymbol{\mathcal{H}}_{xx,T}$ and $\boldsymbol{W}_{T}=\boldsymbol{\mathcal{H}}_{x\theta,T}$. Here, $\boldsymbol{A}_{t}=\boldsymbol{F}_{t}-\boldsymbol{G}_{t}(\boldsymbol{\mathcal{H}}_{uu,t})^{-1}\boldsymbol{\mathcal{H}}_{ux,t}$, $\boldsymbol{B}_{t}=\boldsymbol{G}_{t}(\boldsymbol{\mathcal{H}}_{uu,t})^{-1}\boldsymbol{G}_{t}^\prime$, $\boldsymbol{M}_{t}=\boldsymbol{E}_{t}-\boldsymbol{G}_{t}(\boldsymbol{\mathcal{H}}_{uu,t})^\prime \boldsymbol{\mathcal{H}}_{u\theta,t}$, $\boldsymbol{C}_{t}=\boldsymbol{\mathcal{H}}_{xx,t}^{}-\boldsymbol{\mathcal{H}}_{xu,t}(\boldsymbol{\mathcal{H}}_{uu,t})^{-1}\boldsymbol{\mathcal{H}}_{ux,t}$, $\boldsymbol{N}_{t}=\boldsymbol{\mathcal{H}}_{x\theta,t}-\boldsymbol{\mathcal{H}}_{xu,t}(\boldsymbol{\mathcal{H}}_{uu,t})^\prime \boldsymbol{\mathcal{H}}_{u\theta,t}$ are all known. 
Then, $\frac{\partial \boldsymbol{x}({\boldsymbol{\theta}})}{\partial \boldsymbol{\theta}}$ is obtained by recursively solving the following equations from $t=\hat{t}$ to $T-1$ with $\boldsymbol{X}_{\hat{t}}(\boldsymbol{\theta})=\boldsymbol{0}$:
\begin{equation*} \label{eq:recursiveGG}
\begin{aligned}
&\boldsymbol{U}_{t}=-\boldsymbol{\mathcal{H}}_{uu,t}^{-1}(\boldsymbol{\mathcal{H}}_{ux,t}\boldsymbol{X}_{t}+\boldsymbol{\mathcal{H}}_{u\theta,t}+\boldsymbol{G}_{t}^\prime(\boldsymbol{I}+\boldsymbol{V}_{t+1}\boldsymbol{B}_{t})^{\text{-}1} \boldsymbol{\Gamma}_t ), \\
&\boldsymbol{X}_{t+1}=\boldsymbol{F}_{t}\boldsymbol{X}_{t}+\boldsymbol{G}_{t}\boldsymbol{U}_{t}+\boldsymbol{E}_{t}, \\
& \boldsymbol{\Gamma}_t \triangleq \boldsymbol{V}_{t+1}\boldsymbol{A}_{t}\boldsymbol{X}_{t}+\boldsymbol{V}_{t+1}\boldsymbol{M}_{t}+\boldsymbol{W}_{t+1}, \\
&\boldsymbol{X}_{t}\triangleq \textstyle\frac{\partial\boldsymbol{x}_{t}({\boldsymbol{\theta}})}{\partial\boldsymbol{\theta}}\in\mathbb{R}^{n\times s}, \  \boldsymbol{U}_{t}\triangleq \textstyle\frac{\partial\boldsymbol{u}_{t}({\boldsymbol{\theta}})}{\partial\boldsymbol{\theta}}\in\mathbb{R}^{m\times s}.
\end{aligned}
\end{equation*}
\end{lemma}

With the online update law \eqref{eq:EKF} and gradient obtained from Lemma \ref{lemma:PDP}, the main algorithm is summarized as follows.

\begin{algorithm}
\caption{Online Intention Prediction} \label{algorithm:main}
\textbf{Initialize: } $\hat{\boldsymbol{x}}_{\mathrm{g},0},\hat{\boldsymbol{x}}_{0},\hat{\boldsymbol{\theta}}_{0}, \boldsymbol{P}_{0},\boldsymbol{R}_{t},\boldsymbol{\mathcal{X}}=\{\boldsymbol{x}_0^*\},T_{\mathrm{m}}$ \\
\For{$t=1,...,T$}{
    $\hat{t}= \max(t-T_{\mathrm{m}},0)$\\
    Obtain new observation $\bar{\boldsymbol{x}}_t^*$\\
    Update memory $\boldsymbol{\mathcal{X}}=\{\bar{\boldsymbol{x}}_{\hat{t}}^*,\cdots,\bar{\boldsymbol{x}}_t^*\}$\\
    Propagate $\hat{\boldsymbol{x}}_{\hat{t}+1:T}, \hat{\boldsymbol{u}}_{\hat{t}:T-1}$ by solving \eqref{eq:est_auto_sys} w/ $\hat{\boldsymbol{\theta}}_{t-1}$\\
    Obtain $\boldsymbol{H}_t$ using Lemma \ref{lemma:PDP} with $\hat{\boldsymbol{x}}_t(\hat{\boldsymbol{\theta}}_{t-1})$\\
    Update $\hat{\boldsymbol{\theta}}_t$, $\boldsymbol{K}_t$, $\boldsymbol{P}_t$ using \eqref{eq:EKF}\\
    Extract $\hat{\boldsymbol{x}}_{\mathrm{g},t}$ from $\hat{\boldsymbol{\theta}}_t$\\
}
\end{algorithm}
For each time index \(t=1,\dots,T\), the propagation start time \(\hat{t}\) is determined, and the memory buffer \(\boldsymbol{\mathcal{X}}\) is updated with the new observation \(\bar{\boldsymbol{x}}_t^*\) and \(\hat{t}\).
The predicted trajectory is then propagated by solving the optimal control problem \eqref{eq:est_auto_sys} using the current parameter estimate and the propagation start state \(\bar{\boldsymbol{x}}_{\hat{t}}^*\).
The gradient \(\boldsymbol{H}_t\) is computed from Lemma~\ref{lemma:PDP} based on the estimated state \(\hat{\boldsymbol{x}}_t(\hat{\boldsymbol{\theta}}_{t-1})\). 
After updating the parameters according to \eqref{eq:EKF}, the predicted goal \(\hat{\boldsymbol{x}}_{\mathrm{g},t}\) is obtained from the updated parameter vector \(\hat{\boldsymbol{\theta}}_t\).

\section{Numerical Experiments}\label{sec:simulation}

This section demonstrates the capability of the proposed online intention prediction framework with various experiments performed on a quadrotor drone. The detailed system dynamics are provided in \cite{liang2024online}. 
The unknown parameters in system dynamics are the mass, moment of inertia, length of the wing, and torque coefficient.
The system state vector is $\boldsymbol{x}\triangleq \text{col}\{\boldsymbol{p}_I,\boldsymbol{v}_I,\boldsymbol{q}_{B/I},\boldsymbol{\omega}_I\}\in\mathbb{R}^{13}$ where $\boldsymbol{p}_I\in\mathbb{R}^3$  and $\boldsymbol{v}_I\in\mathbb{R}^3$ are the position and velocity vector of the drone; $\boldsymbol\omega_B\in\mathbb{R}^3$ is the angular velocity vector of the drone; $\boldsymbol{q}_{B/I}\in\mathbb{R}^4$ is the unit quaternion that describes the drone's attitude with respect to the inertial frame. The control inputs $\boldsymbol{u}\in\mathbb{R}^4$ are the thrust from four propellers. The running cost and final cost are $\boldsymbol{\omega}_{\mathrm{r}}^{\prime} c_{\mathrm{r}}(\boldsymbol{x}_t,\boldsymbol{u}_t,\boldsymbol{x}_{\mathrm{g}}) =\boldsymbol{\omega}_{\mathrm{r}}^{\prime}||\boldsymbol{u}_t||^2$ and $\boldsymbol{\omega}_{\mathrm{f}}^{\prime} c_{\mathrm{f}}(\boldsymbol{x}_T,\boldsymbol{x}_{\mathrm{g}}) = \boldsymbol{\omega}_{\mathrm{f}}^{\prime}||\boldsymbol{x}_T-\boldsymbol{x}_{\mathrm{g}}||^2$, respectively.

Experiments are conducted under various noise levels, with 100 random trials for each case. 
In all experiments, the goal state is randomly generated but kept fixed (no goal switching). 
The initial guesses for the system parameters are randomized within \(\pm 25\%\) of their true values, and the system’s initial state is used as the initial prediction. Readers are referred to \cite{liang2024online} for details of the covariance initialization. In this work, the covariance associated with the intention variable is initialized to a larger value than that of the other parameters.

\begin{figure}
\centering
\subfloat[Prediction Loss]
{\label{fig:prediction_loss}
\includegraphics[width=0.48\linewidth]{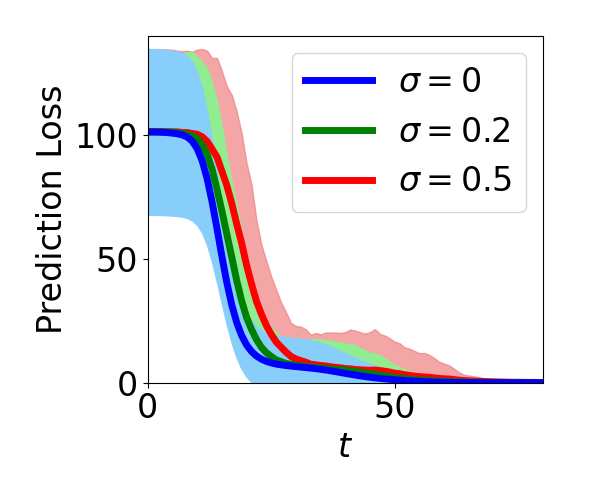}}%
\subfloat[Position]
{\label{fig:state}
\includegraphics[width=0.48\linewidth]{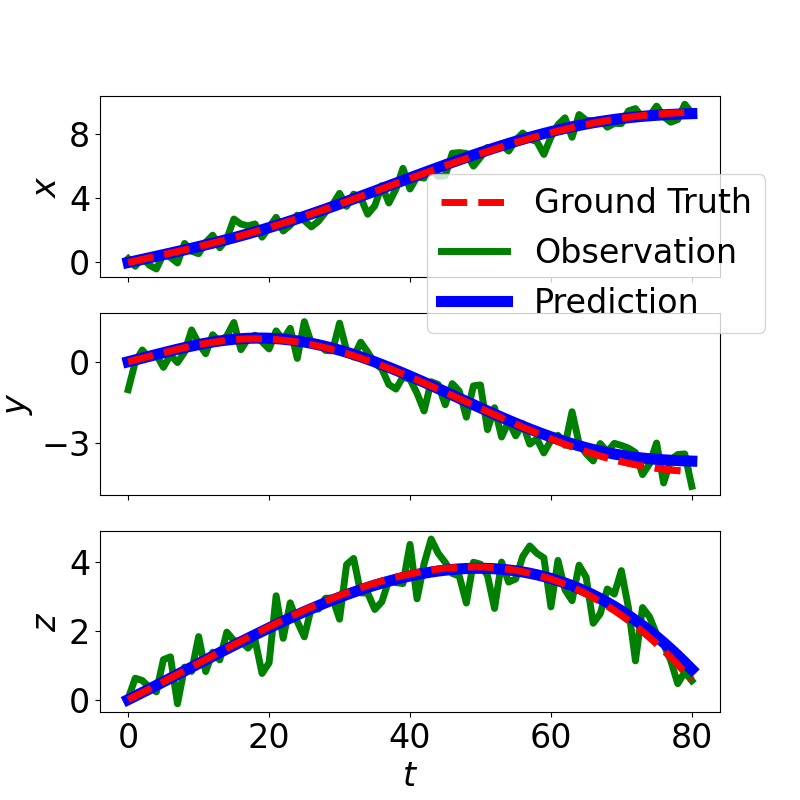}}%
\caption{Intention prediction for quadrotor. (No goal change)}
\label{fig:simulation}
\end{figure} 

Fig.~\ref{fig:prediction_loss} presents the prediction loss against different levels of Gaussian noise, modeled as 
\(\mathcal{N}(\boldsymbol{x}_t^*, \sigma^2\boldsymbol{I}_n)\). The solid lines indicate the average loss, and the shaded regions represent the range of three standard deviations. The prediction loss at each time index \(t\) is computed as 
\(\lVert\hat{\boldsymbol{x}}_{\mathrm{g},t}-\boldsymbol{x}_{\mathrm{g},t}^*\rVert^2\). 
These results demonstrate fast and accurate prediction of the goal. 
Fig.~\ref{fig:state} illustrates the predicted trajectory under a high noise level of  \(\sigma=0.5\), highlighting the method’s robustness to significant measurement noise.
The average prediction time on an Intel i9-14900K CPU was \(60\pm4\) ms per update with a time step of 150 ms, demonstrating that the proposed algorithm is computationally efficient and suitable for real-time usage.

\begin{figure*}
\centering
\subfloat[Prediction Loss]
{\label{fig:switch_loss}\includegraphics[width=0.28\linewidth]{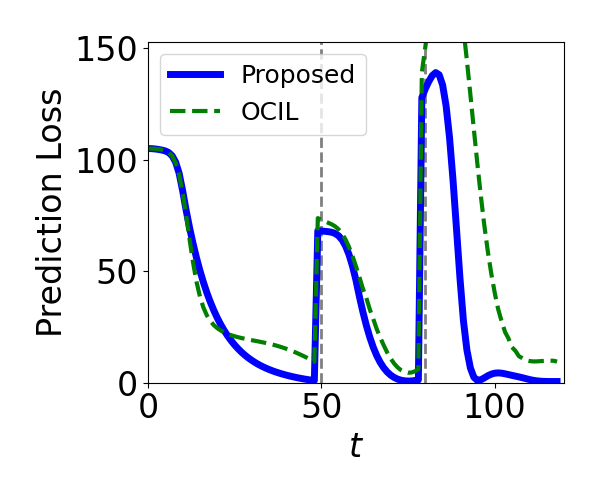}}
\hfill
\subfloat[Trajectory at three representative time frames]
{\label{fig:switch_demo}\includegraphics[width=0.7\linewidth]{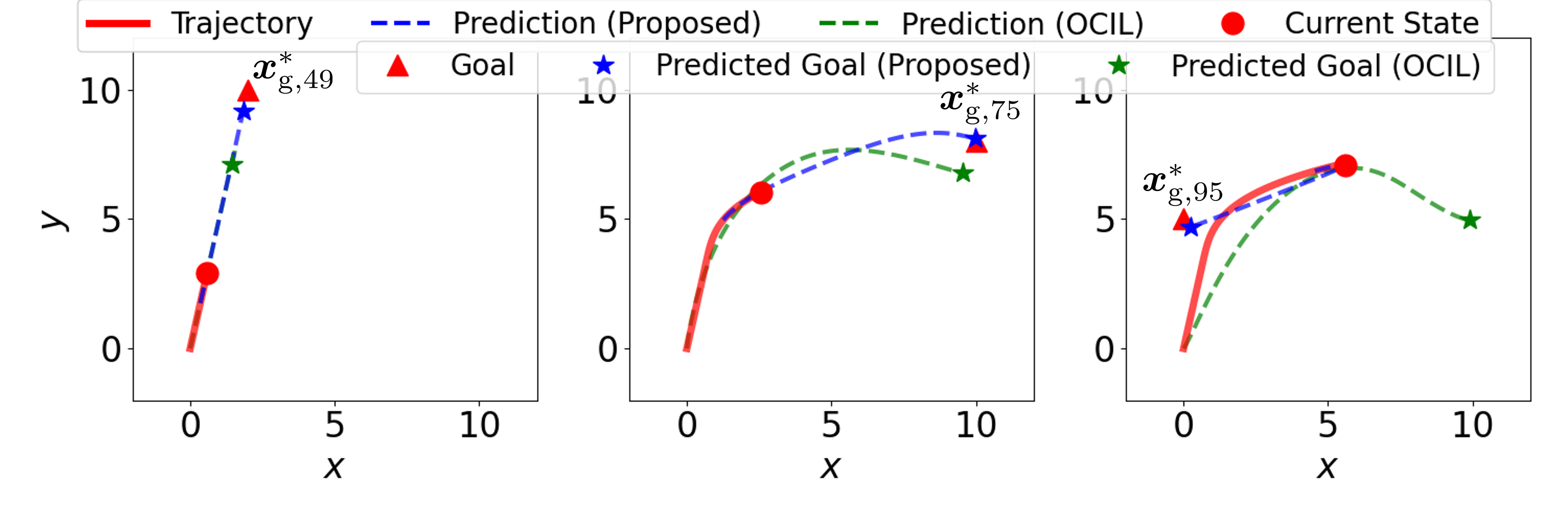}}
\caption{Online intention prediction for quadrotor. (Two goal switches. Vertical dashed lines in (a) indicate goal changes.)}
\label{fig:switch_target}
\end{figure*}

\begin{figure}
\centering
\includegraphics[width=\linewidth]{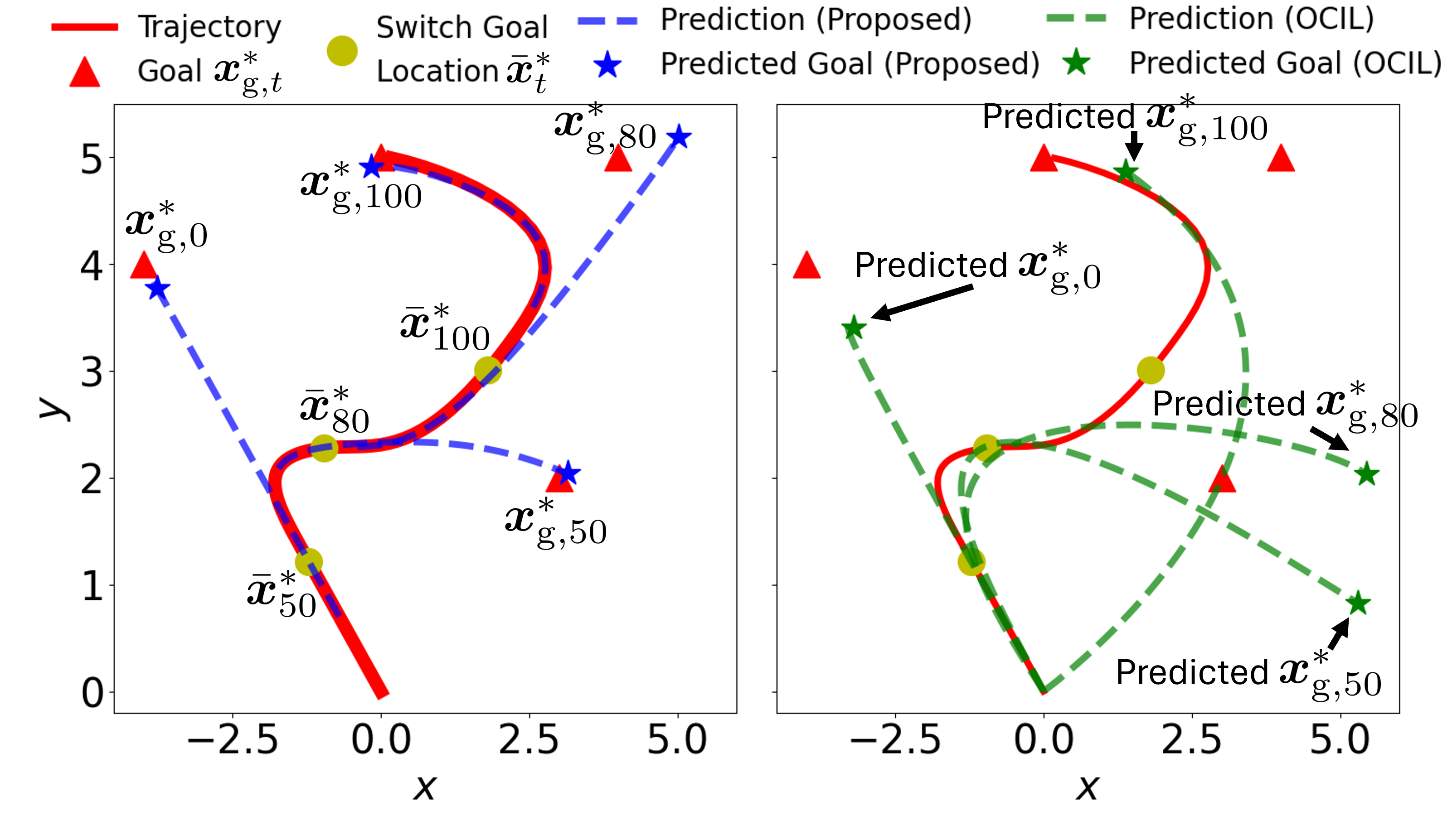}
\caption{Online intention prediction with three goal switches, i.e. at $t=50$, the system is at $\bar{\boldsymbol{x}}_{50}^*$ and switches its goal from $\boldsymbol{x}_{\mathrm{g},0}^*$ to $\boldsymbol{x}_{\mathrm{g},50}^*$ (goal $\boldsymbol{x}_{\mathrm{g},0}^*$, $\cdots$, $\boldsymbol{x}_{\mathrm{g},49}^*$ maintains the same).}
\label{fig:switch_2}
\end{figure}

Furthermore, examples of switching intention are demonstrated. 
Figure~\ref{fig:switch_loss} shows the prediction loss over time when the intention switches twice. 
Distinct peaks appear at each goal switch, followed by a rapid decrease in loss after each change. 
For comparison, we also show the performance of OCIL without the proposed shifting-horizon strategy; in this case the loss fails to converge under switching goals. 
Figure~\ref{fig:switch_demo} presents top-view snapshots of the trajectory at key time frames, illustrating the benefit of the shifting-horizon approach. 
OCIL cannot handle switching targets effectively because its predictions are heavily influenced by outdated observations under the previous goal. Fig.~\ref{fig:switch_2} shows a case with three goal switches. 
The proposed method accurately predicts each goal, whereas OCIL fails and produces poor trajectories because it cannot correctly handle a trajectory composed of segments from different objectives.

\begin{figure}
\centering
\subfloat[]
{\label{fig:compare_loss_traj}\includegraphics[width=0.50\linewidth]{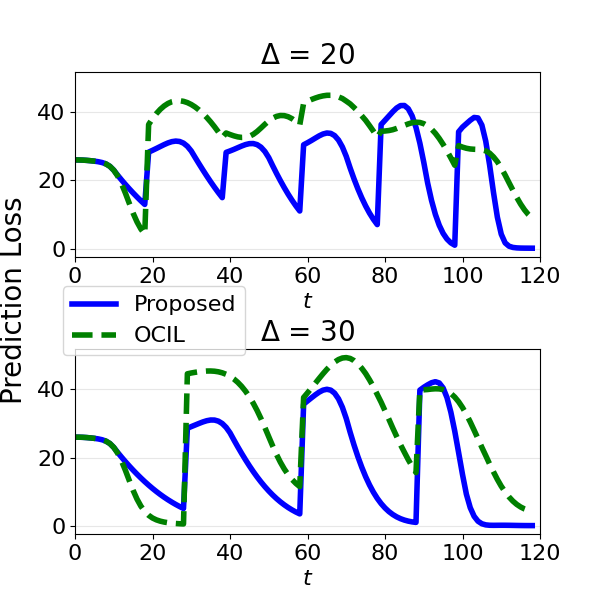}}
\hfill
\subfloat[]
{\label{fig:avg_prediction}\includegraphics[width=0.50\linewidth]{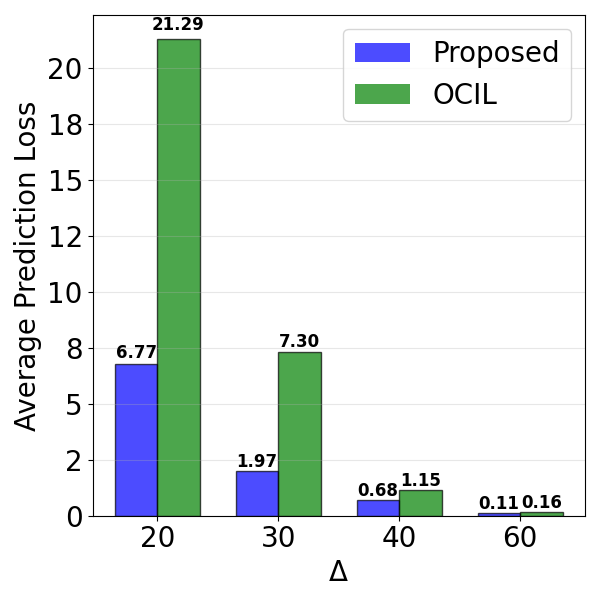}}
\caption{(a) Comparison of prediction loss over time between the proposed method and OCIL. $\Delta$ denotes the time interval between goal switches. (b) Comparison of average prediction loss between the proposed method and OCIL.}
\label{}
\end{figure}

Fig.~\ref{fig:compare_loss_traj} compares the prediction loss over time between the proposed method and OCIL for two scenarios with goal switches, using 20-step and 30-step intervals between successive goal changes.
The proposed method achieves a smaller loss after each switch than OCIL, benefiting from the shifting-horizon strategy. 
Fig.~\ref{fig:avg_prediction} shows the average prediction loss across 100 random cases for each switching interval. 
These results demonstrate that the shifting-horizon strategy consistently yields smaller prediction losses. 
When $\Delta$ is small, both the proposed method and OCIL struggle to reach a small prediction loss due to insufficient time for updating, yet the proposed method still maintains a significantly lower loss than OCIL.

\section{Hardware Experiments} \label{section:hardware}

\begin{figure*}
\centering
\subfloat[]
{\label{fig:uavT35}\includegraphics[width=0.32\linewidth]{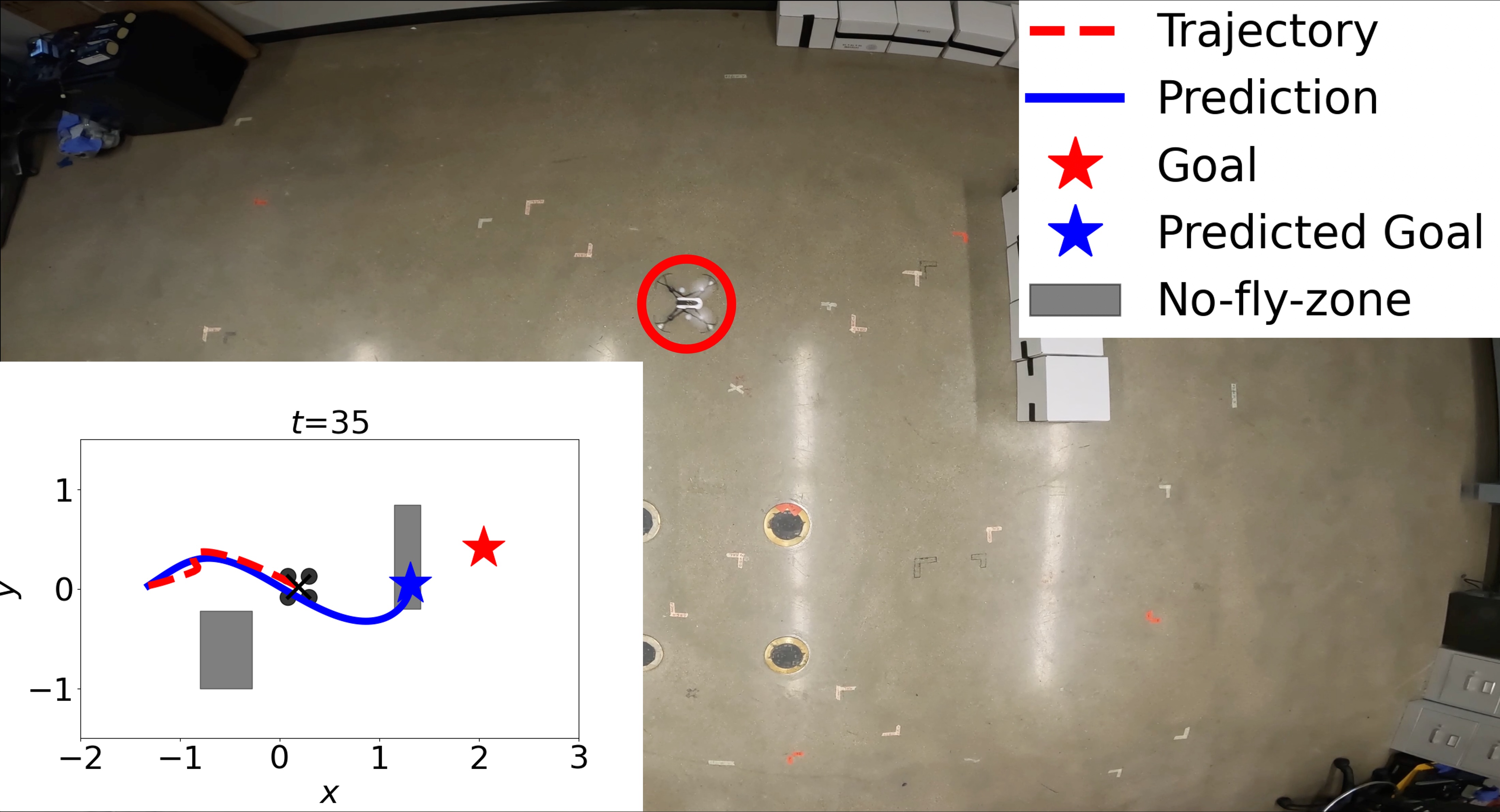}}
\hfill
\subfloat[]
{\label{fig:uavT60}\includegraphics[width=0.32\linewidth]{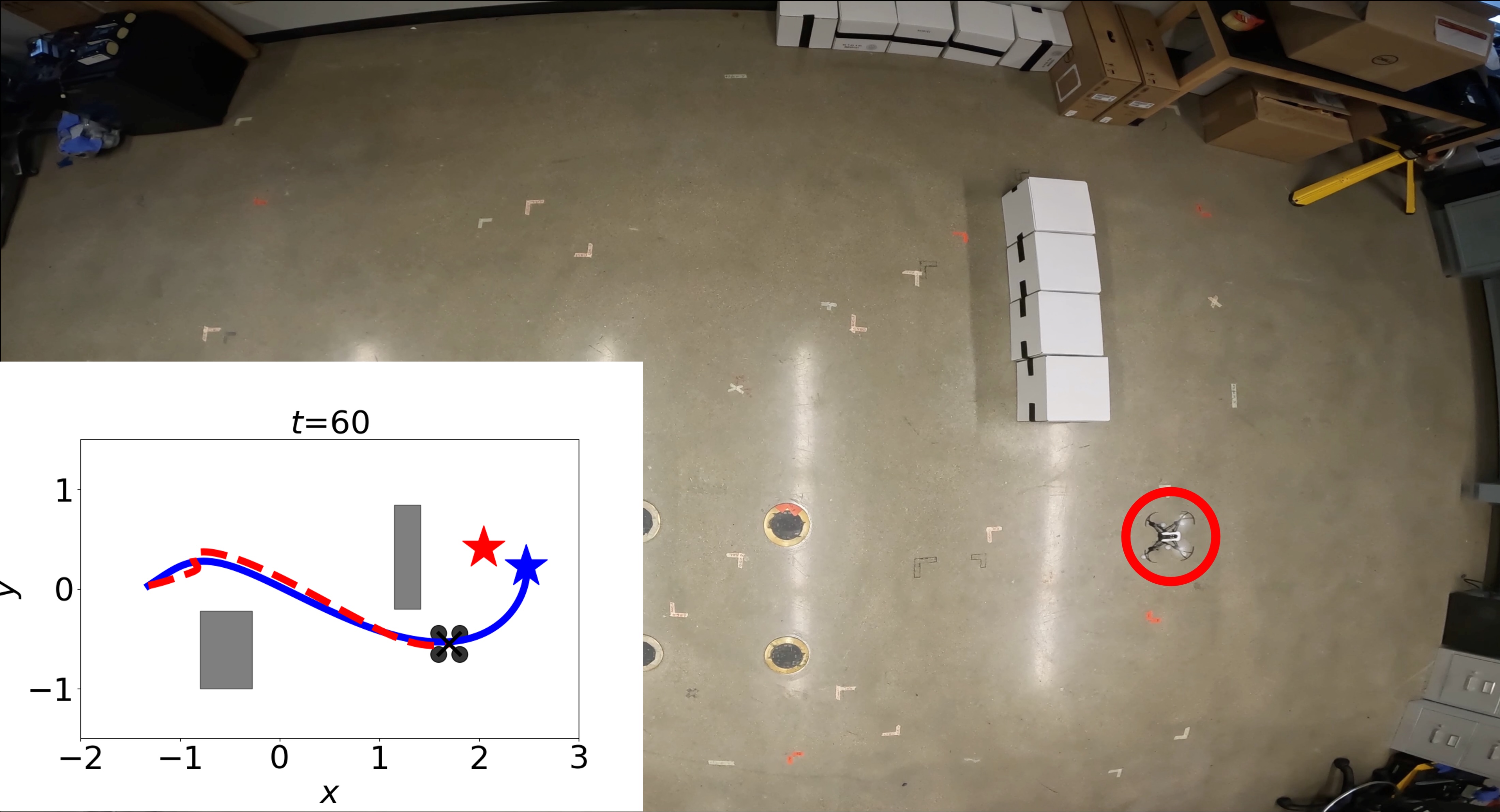}}
\hfill
\subfloat[]
{\label{fig:uavT80}\includegraphics[width=0.32\linewidth]{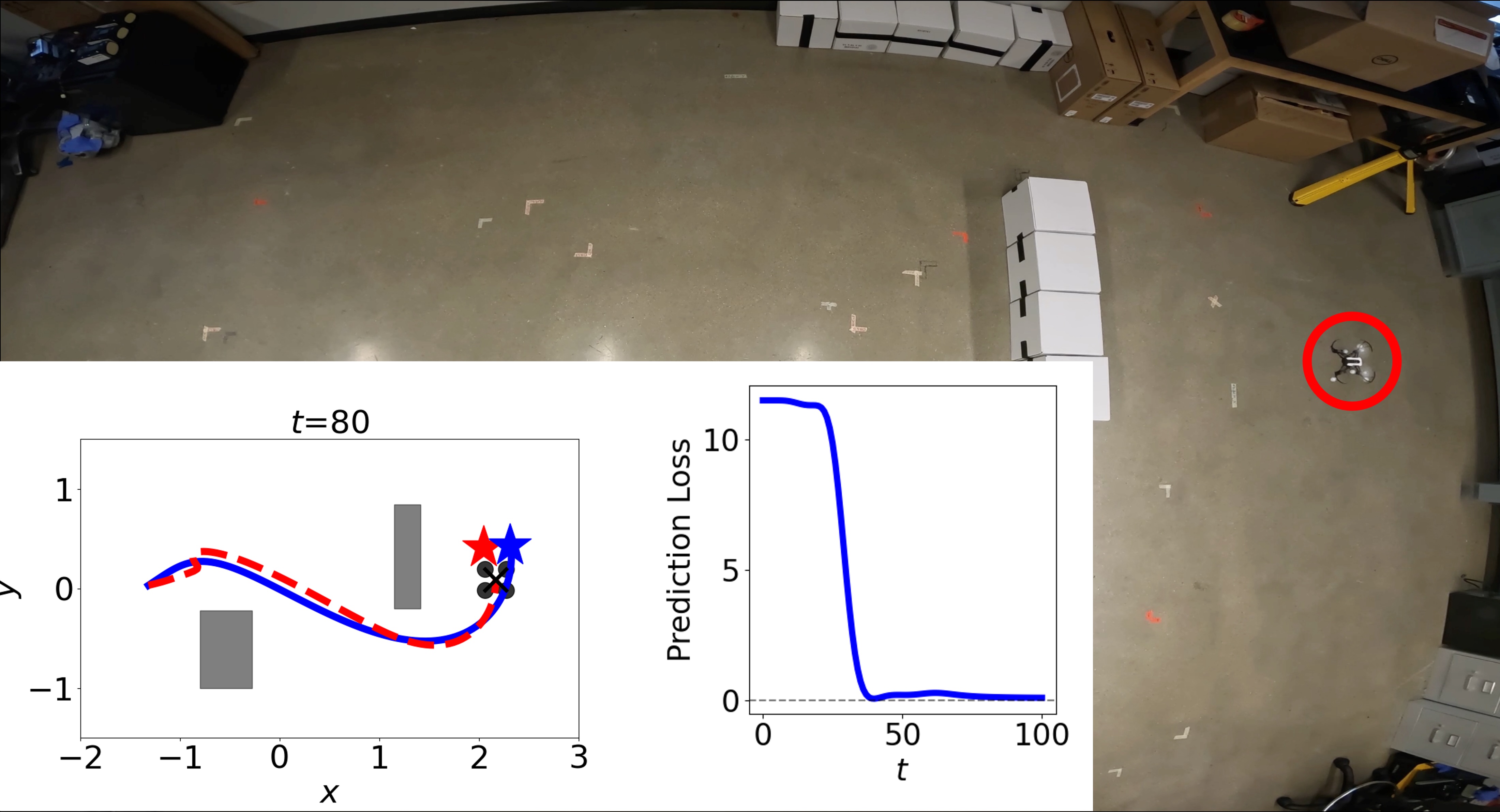}}
\caption{Snapshots of the experiment for a quadrotor drone.
Fig.~\ref{fig:uavT35} shows that at $t=35$, the prediction indicates the quadrotor intends to fly in the positive $x$-direction but does not yet reflect an intention to avoid the no-fly zone or accurately predict its true goal. By $t=60$, as shown in Fig.~\ref{fig:uavT60}, the predicted goal shifts into the correct region as the quadrotor begins its second turn. As more measurements are collected, the prediction continues to improve and eventually converges to the ground truth, as illustrated in Fig.~\ref{fig:uavT80}. The prediction loss is included in Fig.~\ref{fig:uavT80}.}
\label{fig:hardware_exp}
\end{figure*}

Hardware experiments on a quadrotor drone demonstrate the capability of the proposed algorithm on a real-world robotic platform \footnote{Video can be found at https://youtu.be/rKf8zJNEKc8}. The experiments are carried out within a 5 m $\times$ 3 m area, with a motion capture system employed to measure the drone's state.
The running cost and final cost are $
\boldsymbol{\omega}_{\mathrm{r}}^{\prime} c_{\mathrm{r}}(\boldsymbol{x}_t,\boldsymbol{u}_t)  = 
0.1||\boldsymbol{u}_t||^2 + 
100||\boldsymbol{x}_{t}[3] - 0.6||^2 + \boldsymbol{\omega}_{\mathrm{r},\mathrm{obs}}^{\prime}||\boldsymbol{x}_t[1:2]-\boldsymbol{x}_{\mathrm{obs}}[1:2]||^2$ and $
\boldsymbol{\omega}_{\mathrm{f}}^{\prime} c_{\mathrm{f}}(\boldsymbol{x}_T,\boldsymbol{x}_{\mathrm{g}}) = \boldsymbol{\omega}_{\mathrm{f}}^{\prime}||\boldsymbol{x}_T-\boldsymbol{x}_{\mathrm{g}}||^2$, respectively.
Here, $\boldsymbol{\omega}_{\mathrm{r}} =\boldsymbol{\omega}_{\mathrm{r},\mathrm{obs}} $ and $\boldsymbol{x}_{\mathrm{obs}}[1:2]=\matt{0 & 5}^\prime$. The goal state is $\boldsymbol{x}_{\mathrm{g}}^{*}=\matt{2 & 0 & 0.6 & \boldsymbol{0}_{1\times6} & 1 & \boldsymbol{0}_{1\times3}}^\prime$.


Fig.~\ref{fig:hardware_exp} illustrates the prediction of the quadrotor drone at selected times.
The average prediction time per step is 92.3 ms, which is lower than the motion capture system's measuring time step 100 ms. The prediction loss slightly increases during the second turn but continues to converge toward zero as more observations are collected.

\section{Conclusions} \label{section:conclusion}

This paper introduces an online framework for predicting the intention (goal state) of autonomous systems operating in uncertain and dynamic environments. Unlike prior approaches that assume a fixed or limited set of goals, our formulation treats the goal as a parameter and allows it to change during execution, even when the system’s dynamics or objective parameters are unknown. The method casts intention prediction as an IOC / IRL problem and employs a shifting horizon update scheme powered by online control-informed learning to update goal estimates as new data arrive. Experiments with a quadrotor, along with simulations under varying noise levels, confirm that the proposed approach enables fast and accurate real-time intention prediction.



\bibliographystyle{IEEEtran}
\bibliography{zihao, ref_zehui, ref_l4dc}

@article{brannstrom2010model,
  title={Model-based threat assessment for avoiding arbitrary vehicle collisions},
  author={Br{\"a}nnstr{\"o}m, Mattias and Coelingh, Erik and Sj{\"o}berg, Jonas},
  journal={IEEE Trans. Intell. Transp. Syst.},
  volume={11},
  number={3},
  pages={658--669},
  year={2010},
  publisher={IEEE}
}

@article{lefkopoulos2020interaction,
  title={Interaction-aware motion prediction for autonomous driving: A multiple model kalman filtering scheme},
  author={Lefkopoulos, Vasileios and Menner, Marcel and Domahidi, Alexander and Zeilinger, Melanie N},
  journal={IEEE RA-L},
  volume={6},
  number={1},
  pages={80--87},
  year={2020},
  publisher={IEEE}
}

@article{wang2019trajectory,
  title={Trajectory planning and safety assessment of autonomous vehicles based on motion prediction and model predictive control},
  author={Wang, Yijing and Liu, Zhengxuan and Zuo, Zhiqiang and Li, Zheng and Wang, Li and Luo, Xiaoyuan},
  journal={IEEE IEEE Trans. Veh. Technol.},
  volume={68},
  number={9},
  pages={8546--8556},
  year={2019},
  publisher={IEEE}
}

@inproceedings{guo2019modeling,
  title={Modeling multi-vehicle interaction scenarios using gaussian random field},
  author={Guo, Yaohui and Kalidindi, Vinay Varma and Arief, Mansur and Wang, Wenshuo and Zhu, Jiacheng and Peng, Huei and Zhao, Ding},
  booktitle={2019 IEEE ITSC},
  pages={3974--3980},
  year={2019},
  organization={IEEE}
}

@inproceedings{kumar2013learning,
  title={Learning-based approach for online lane change intention prediction},
  author={Kumar, Puneet and Perrollaz, Mathias and Lefevre, St{\'e}phanie and Laugier, Christian},
  booktitle={2013 IEEE Intelligent Vehicles Symposium (IV)},
  pages={797--802},
  year={2013},
  organization={IEEE}
}

@ARTICLE{li2020pedestrian,
  author={Li, Yang and Lu, Xiao-Yun and Wang, Jianqiang and Li, Keqiang},
  journal={IEEE Trans. Intell. Veh.}, 
  title={Pedestrian Trajectory Prediction Combining Probabilistic Reasoning and Sequence Learning}, 
  year={2020},
  volume={5},
  number={3},
  pages={461-474},
  doi={10.1109/TIV.2020.2966117}}

@article{chandra2020forecasting,
  title={Forecasting trajectory and behavior of road-agents using spectral clustering in graph-lstms},
  author={Chandra, Rohan and Guan, Tianrui and Panuganti, Srujan and Mittal, Trisha and Bhattacharya, Uttaran and Bera, Aniket and Manocha, Dinesh},
  journal={IEEE RA-L},
  volume={5},
  number={3},
  pages={4882--4890},
  year={2020},
  publisher={IEEE}
}

@article{wu2020efficient,
  title={Efficient sampling-based maximum entropy inverse reinforcement learning with application to autonomous driving},
  author={Wu, Zheng and Sun, Liting and Zhan, Wei and Yang, Chenyu and Tomizuka, Masayoshi},
  journal={IEEE RA-L},
  volume={5},
  number={4},
  pages={5355--5362},
  year={2020},
  publisher={IEEE}
}

@article{huang2021driving,
  title={Driving behavior modeling using naturalistic human driving data with inverse reinforcement learning},
  author={Huang, Zhiyu and Wu, Jingda and Lv, Chen},
  journal={IEEE Trans. Intell. Transp. Syst.},
  volume={23},
  number={8},
  pages={10239--10251},
  year={2021},
  publisher={IEEE}
}

@article{boutayeb2002convergence,
  title={Convergence analysis of the extended Kalman filter used as an observer for nonlinear deterministic discrete-time systems},
  author={Boutayeb, Mohamed and Rafaralahy, Hugues and Darouach, Mohamed},
  journal={IEEE Transactions on Automatic Control},
  volume={42},
  number={4},
  pages={581--586},
  year={2002},
  publisher={IEEE}
}

@inproceedings{jin2020pontryagin,
  title={Pontryagin differentiable programming: An end-to-end learning and control framework},
  author={Jin, Wanxin and Wang, Zhaoran and Yang, Zhuoran and Mou, Shaoshuai},
  booktitle={NeurIPS},
  pages={7979--7992},
  year={2020}
}

@string{ICML  = {International Conference on Machine Learning}}

@inproceedings{ziebart2008maximum,
	title={Maximum entropy inverse reinforcement learning},
	author={Ziebart, Brian D and Maas, Andrew and Bagnell, J Andrew and Dey, Anind K},
	booktitle={AAAI Conference on Artificial Intelligence},
	pages={1433--1438},
	year={2008}
}

@inproceedings{abbeel2004apprenticeship,
	title={Apprenticeship learning via inverse reinforcement learning},
	author={Abbeel, Pieter and Ng, Andrew Y},
	booktitle={ICML},
	pages={1--8},
	year={2004}
}

@INPROCEEDINGS{liang2023data,
  author={Liang, Zihao and Hao, Wenjian and Mou, Shaoshuai},
  booktitle={2023 IEEE CDC}, 
  title={A Data-Driven Approach for Inverse Optimal Control}, 
  year={2023},
  volume={},
  number={},
  pages={3632-3637},
  doi={10.1109/CDC49753.2023.10383220}}

@INPROCEEDINGS{liang2022iterative,
  author={Liang, Zihao and Jin, Wanxin and Mou, Shaoshuai},
  booktitle={2022 ASCC}, 
  title={An Iterative Method for Inverse Optimal Control}, 
  year={2022},
  pages={959-964},
  doi={10.23919/ASCC56756.2022.9828009}}

@article{liang2024online,
  title={Online Control-Informed Learning},
  author={Liang, Zihao and Zhou, Tianyu and Lu, Zehui and Mou, Shaoshuai},
  journal={Transactions on Machine Learning Research},
  year={2025}
}

@inproceedings{wang2019imitation,
  title={Imitation learning for human pose prediction},
  author={Wang, Borui and Adeli, Ehsan and Chiu, Hsu-kuang and Huang, De-An and Niebles, Juan Carlos},
  booktitle={Proceedings of the IEEE/CVF ICCV},
  pages={7124--7133},
  year={2019}
}

@inproceedings{macglashan2015between,
  title={Between Imitation and Intention Learning.},
  author={MacGlashan, James and Littman, Michael L},
  booktitle={IJCAI},
  volume={15},
  pages={3692--3698},
  year={2015}
}

@article{mcghan2015human,
  title={Human intent prediction using markov decision processes},
  author={McGhan, Catharine LR and Nasir, Ali and Atkins, Ella M},
  journal={Journal of Aerospace Information Systems},
  volume={12},
  number={5},
  pages={393--397},
  year={2015},
  publisher={American Institute of Aeronautics and Astronautics}
}

@article{liu2020spatiotemporal,
  title={Spatiotemporal relationship reasoning for pedestrian intent prediction},
  author={Liu, Bingbin and Adeli, Ehsan and Cao, Zhangjie and Lee, Kuan-Hui and Shenoi, Abhijeet and Gaidon, Adrien and Niebles, Juan Carlos},
  journal={IEEE RA-L},
  volume={5},
  number={2},
  pages={3485--3492},
  year={2020},
  publisher={IEEE}
}

@article{rudenko2020human,
  title={Human motion trajectory prediction: A survey},
  author={Rudenko, Andrey and Palmieri, Luigi and Herman, Michael and Kitani, Kris M and Gavrila, Dariu M and Arras, Kai O},
  journal={Int. J. Robot. Res.},
  volume={39},
  number={8},
  pages={895--935},
  year={2020},
  publisher={Sage Publications Sage UK: London, England}
}

@article{huang2022survey,
  title={A survey on trajectory-prediction methods for autonomous driving},
  author={Huang, Yanjun and Du, Jiatong and Yang, Ziru and Zhou, Zewei and Zhang, Lin and Chen, Hong},
  journal={IEEE Trans. Intell. Veh.},
  volume={7},
  number={3},
  pages={652--674},
  year={2022},
  publisher={IEEE}
}

@inproceedings{gu2021densetnt,
  title={Densetnt: End-to-end trajectory prediction from dense goal sets},
  author={Gu, Junru and Sun, Chen and Zhao, Hang},
  booktitle={Proceedings of the IEEE/CVF ICCV},
  pages={15303--15312},
  year={2021}
}

@inproceedings{hu2022model,
  title={Model-based imitation learning for urban driving},
  author={Hu, Anthony and Corrado, Gianluca and Griffiths, Nicolas and Murez, Zachary and Gurau, Corina and Yeo, Hudson and Kendall, Alex and Cipolla, Roberto and Shotton, Jamie},
  booktitle={NeurIPS},
  pages={20703--20716},
  year={2022}
}

@article{duan2024structured,
  title={A structured prediction approach for robot imitation learning},
  author={Duan, Anqing and Batzianoulis, Iason and Camoriano, Raffaello and Rosasco, Lorenzo and Pucci, Daniele and Billard, Aude},
  journal={Int. J. Robot. Res.},
  volume={43},
  number={2},
  pages={113--133},
  year={2024},
  publisher={Sage Publications Sage UK: London, England}
}

@inproceedings{zhao2021tnt,
  title={Tnt: Target-driven trajectory prediction},
  author={Zhao, Hang and Gao, Jiyang and Lan, Tian and Sun, Chen and Sapp, Ben and Varadarajan, Balakrishnan and Shen, Yue and Shen, Yi and Chai, Yuning and Schmid, Cordelia and others},
  booktitle={CoRL},
  pages={895--904},
  year={2021},
  organization={PMLR}
}

@inproceedings{elnaggar2018irl,
  title={An IRL approach for cyber-physical attack intention prediction and recovery},
  author={Elnaggar, Mahmoud and Bezzo, Nicola},
  booktitle={2018 Annual American Control Conference (ACC)},
  pages={222--227},
  year={2018},
  organization={IEEE}
}

@inproceedings{best2015bayesian,
  title={Bayesian intention inference for trajectory prediction with an unknown goal destination},
  author={Best, Graeme and Fitch, Robert},
  booktitle={2015 IEEE/RSJ IROS},
  pages={5817--5823},
  year={2015},
  organization={IEEE}
}

@ARTICLE{zhou2025safe,
  author={Zhou, Tianyu and Liang, Zihao and Lu, Zehui and Mou, Shaoshuai},
  journal={IEEE Control Systems Letters}, 
  title={Safe Online Control-Informed Learning}, 
  year={2025},
  volume={9},
  number={},
  pages={3083-3088},
  }

\end{document}